\documentclass[letterpaper]{article} 
\usepackage{aaai25}  
\usepackage{times}  
\usepackage{helvet}  
\usepackage{courier}  
\usepackage[hyphens]{url}  
\usepackage{graphicx} 
\usepackage{natbib}  
\usepackage{caption} 
\usepackage{algorithm}
\usepackage{algorithmic}

\usepackage{multirow}
\usepackage{tabularx}
\usepackage{array}
\usepackage{subcaption}
\usepackage{xcolor}
\usepackage{amsmath}
\usepackage{amssymb}
\usepackage{booktabs}

\usepackage{newfloat}
\usepackage{listings}
\DeclareCaptionStyle{ruled}{labelfont=normalfont,labelsep=colon,strut=off} 
\floatstyle{ruled}
\newfloat{listing}{tb}{lst}{}
\floatname{listing}{Listing}
\title{\textit{AWRaCLe}: All-Weather Image Restoration using Visual In-Context Learning}
\author{
    Sudarshan Rajagopalan,
    Vishal M. Patel
}
\affiliations{
    Johns Hopkins University

    \{sambasa2, vpatel36\}@jhu.edu
}

\usepackage{bibentry}

\newcommand\blfootnote[1]{%
  \begingroup
  \renewcommand\thefootnote{}\footnote{#1}%
  \addtocounter{footnote}{-1}%
  \endgroup
}
\usepackage{url}
\newcommand{\coloredurl}[2]{\textcolor{#1}{\url{#2}}}

\begin{document}

\maketitle
\blfootnote{Project Page: \coloredurl{blue}{https://sudraj2002.github.io/awraclepage/}}
\begin{abstract}
  All-Weather Image Restoration (AWIR) under adverse weather conditions is a challenging task due to the presence of different types of degradations. Prior research in this domain relies on extensive training data but lacks the utilization of additional contextual information for restoration guidance. Consequently, the performance of existing methods is limited by the degradation cues that are learnt from individual training samples. Recent advancements in visual in-context learning have introduced generalist models that are capable of addressing multiple computer vision tasks simultaneously by using the information present in the provided context as a prior. In this paper, we propose \textit{\textbf{A}ll-\textbf{W}eather Image \textbf{R}estor\textbf{a}tion using Visual In-\textbf{C}ontext \textbf{Le}arning} (AWRaCLe), a novel approach for AWIR that innovatively utilizes degradation-specific visual context information to steer the image restoration process. To achieve this, AWRaCLe incorporates Degradation Context Extraction (DCE) and Context Fusion (CF) to seamlessly integrate degradation-specific features from the context into an image restoration network. The proposed DCE and CF blocks leverage CLIP features and incorporate attention mechanisms to adeptly learn and fuse contextual information. These blocks are specifically designed for visual in-context learning under all-weather conditions and are crucial for effective context utilization. Through extensive experiments, we demonstrate the effectiveness of AWRaCLe for all-weather restoration and show that our method advances the state-of-the-art in AWIR. 
\end{abstract}


\section{Introduction}
\label{sec:introduction}
Unfavorable weather conditions, such as rain, snow and  haze, significantly degrade the performance of computer vision systems impacting applications such as autonomous navigation, surveillance, and aerial imaging. Thus, there is a need for frameworks that mitigate weather-induced corruptions while preserving the underlying image semantics. Intial physics-based methods~\cite{early1,early2,early3} struggled to handle real-world variability in degradations. Deep learning approaches that handle a single degradation~\cite{dehazeformer, snow2, rain2, swinir, restormer, mprnet, uformer} at a time were then proposed, but they must be retrained or fine-tuned for each new condition, reducing their practicality.

\begin{figure*}
    \centering
    \setlength{\tabcolsep}{1pt}
    \begin{tabular}{cccccc}
         \parbox[t]{2mm}{\multirow{2}{*}{\rotatebox[origin=c]{90}{Context Pair\hspace{-15pt}}}}&\includegraphics[height=1.53cm, width=3.1cm]{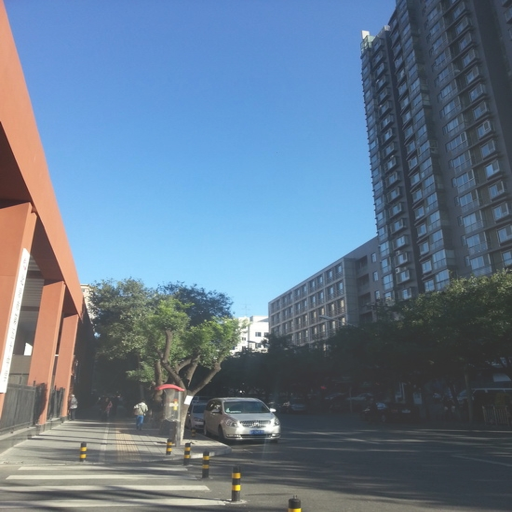}&\includegraphics[height=1.53cm, width=3.1cm]{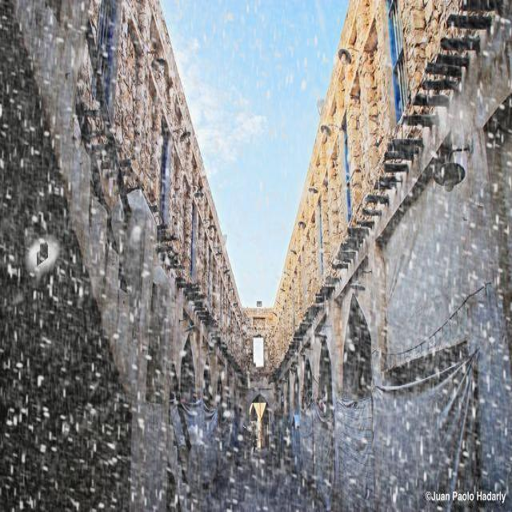}&\includegraphics[height=1.53cm, width=3.1cm]{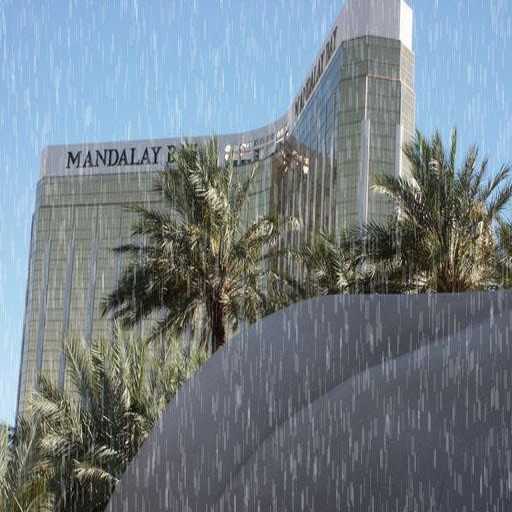}&\includegraphics[height=1.53cm, width=3.1cm]{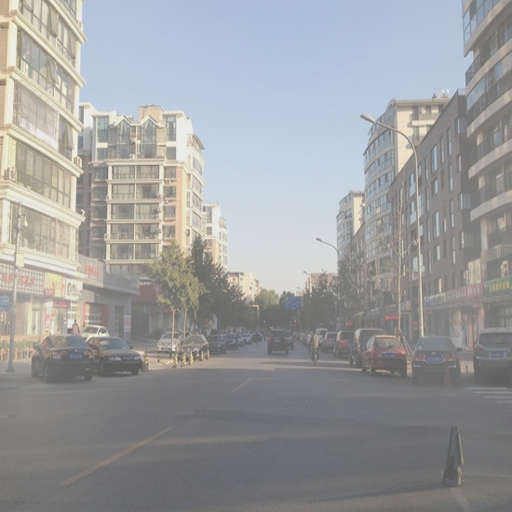}&\includegraphics[height=1.53cm, width=3.1cm]{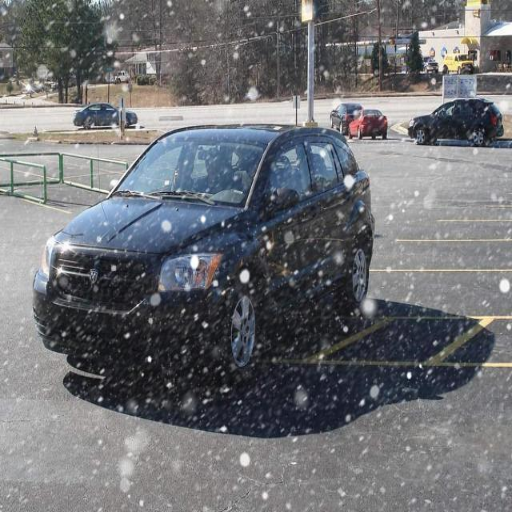}\\
         
         &\includegraphics[height=1.53cm, width=3.1cm]{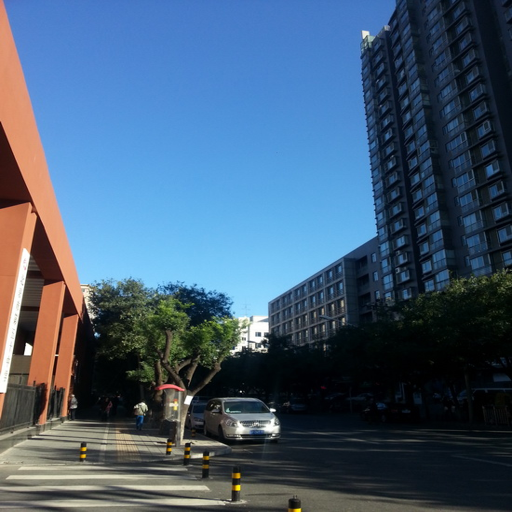}&\includegraphics[height=1.53cm, width=3.1cm]{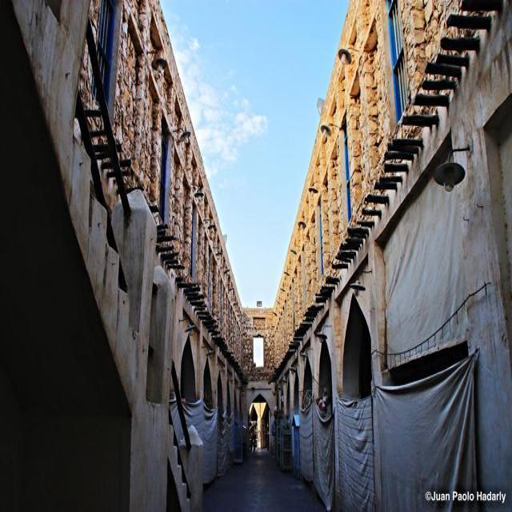}&\includegraphics[height=1.53cm, width=3.1cm]{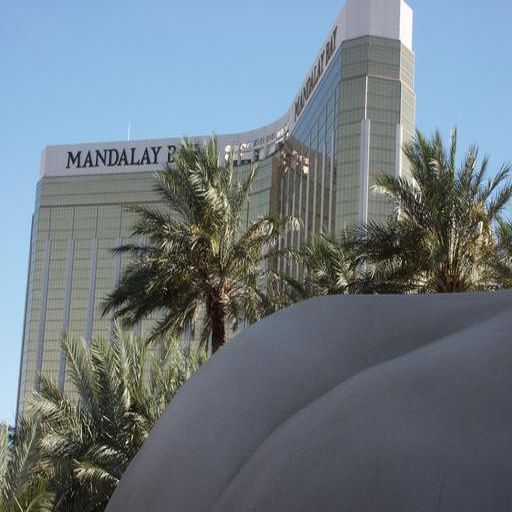}&\includegraphics[height=1.53cm, width=3.1cm]{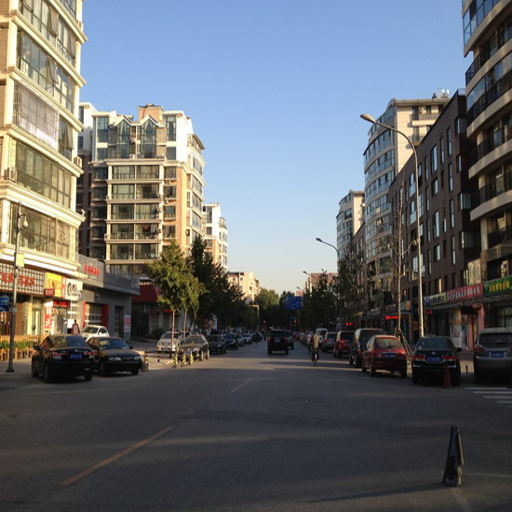}&\includegraphics[height=1.53cm, width=3.1cm]{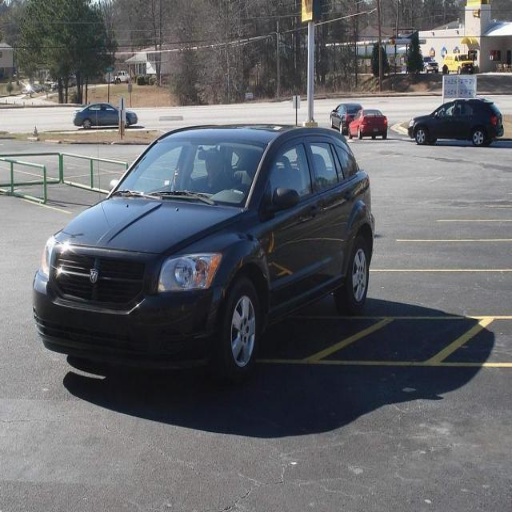}\\

         \hspace{2pt}\rotatebox[origin=c]{90}{Query\hspace{-38pt}}&\includegraphics[height=1.53cm, width=3.1cm]{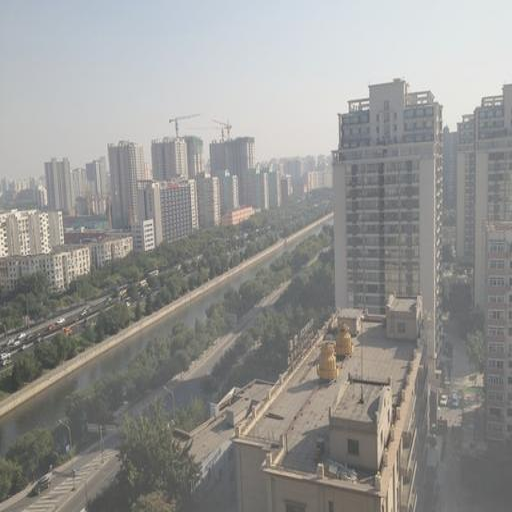}&\includegraphics[height=1.53cm, width=3.1cm]{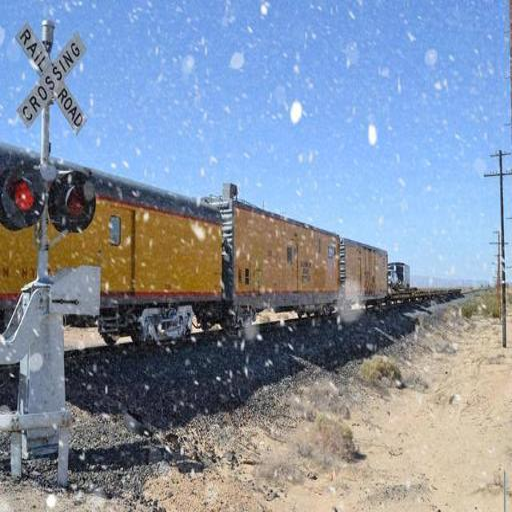}&\includegraphics[height=1.53cm, width=3.1cm]{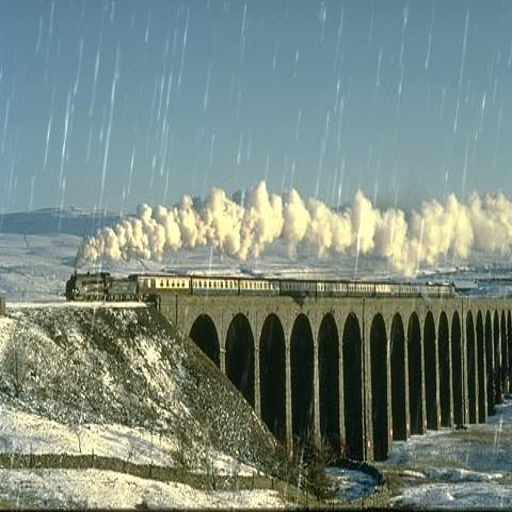}&\includegraphics[height=1.53cm, width=3.1cm]{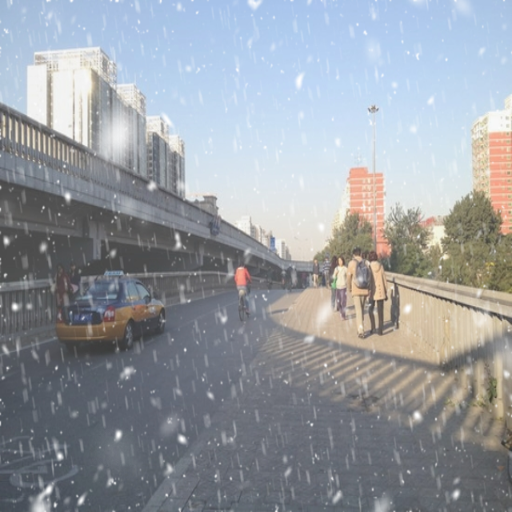}&\includegraphics[height=1.53cm, width=3.1cm]{AnonymousSubmission/fig1_final/inputcsd.png}\\

         \hspace{2pt}\rotatebox[origin=c]{90}{Output\hspace{-38pt}}&\includegraphics[height=1.53cm, width=3.1cm]{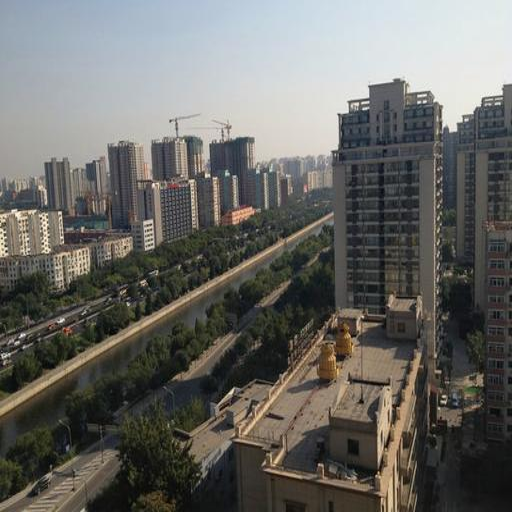}&\includegraphics[height=1.53cm, width=3.1cm]{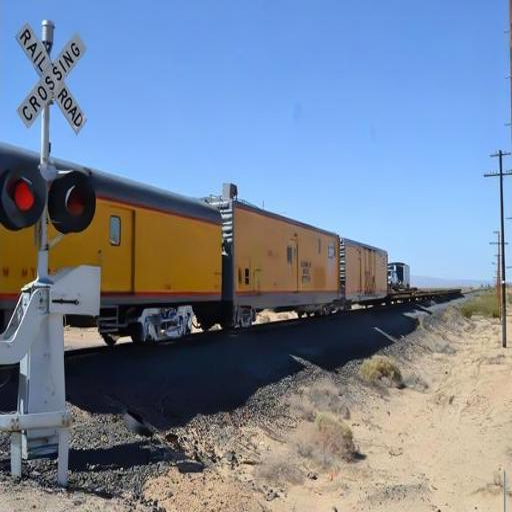}&\includegraphics[height=1.53cm, width=3.1cm]{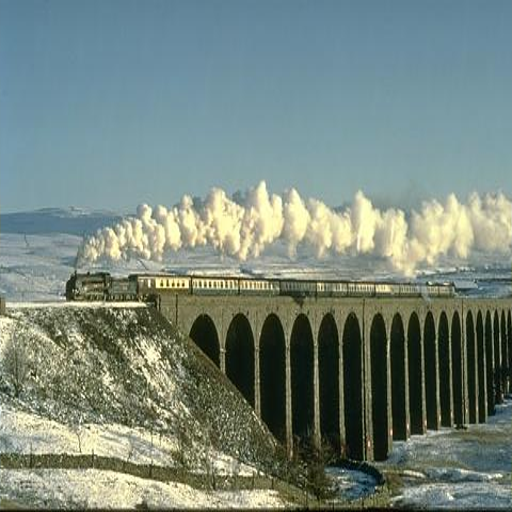}&\includegraphics[height=1.53cm, width=3.1cm]{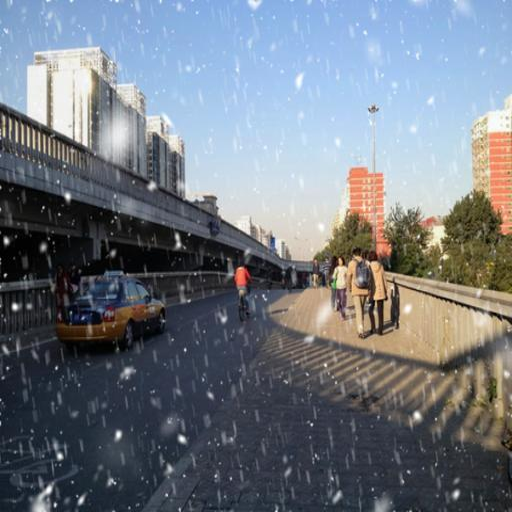}&\includegraphics[height=1.53cm, width=3.1cm]{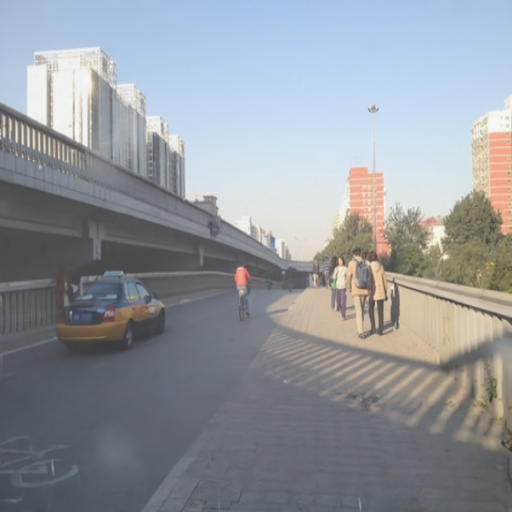}\\
         &(a) Dehaze& (b) Desnow& (c) Derain& (d) Selective Dehaze& (e) Selective Desnow
    \end{tabular}
    \caption{Illustration of AWRaCLe: Our visual in-context learning approach for all-weather image restoration. The first two rows are the context pair. The third row is the query image that needs to be restored and the fourth row is our output. (d) and (e) show results for selective removal of haze and snow, respectively, from an image containing their mixture.}
    \label{fig:first}
\end{figure*}


To address these issues, recent work has focused on All-Weather Image Restoration (AWIR) networks that handle multiple degradations with a single model~\cite{nas,transw,airnet,faig,weatherdiff,tkmc,promptir,diffuir}. However, these approaches learn degradation representations only from individual images and lack guidance that provides detailed degradation-specific information (DSI). This limits their ability to effectively learn features unique to different degradations, impeding their performance. Thus, there is a need for a framework capable of learning robust degradation-specific features that can facilitate effective restoration. This is challenging without supplementary knowledge about the nature of the corruption. While some methods~\cite{textawir,textir} have introduced text-based guidance for AWIR, text descriptions can only convey high-level semantic information about the degradation and fail to describe important aspects of the corruption such as its visual characteristics. 

To address this limitation, recent approaches such as Diff-Plugin~\cite{diffplugin} and DA-CLIP~\cite{daclip} attempt to extract DSI directly from images. Diff-Plugin extracts task-specific and spatial prompts from the degraded input image. However, it requires multiple independently trained task-specific plugins for each degradation. DA-CLIP extracts text-aligned DSI from the degraded image but lacks detailed visual information, as text conveys only high-level features. Moreover, both methods face the challenge of disentangling scene characteristics from DSI because they rely on extracting DSI from a single image. We conjecture that feeding both the clean and degraded images as context to an image restoration network can overcome these limitations as the consistent scene between the pair allows the network to aggregate visual information specific to the degradation. We accomplish this with the help of {\emph{visual in-context learning}}.

In-context learning, as demonstrated by large language models, is very well-studied in Natural Language Processing (NLP). In comparison, visual in-context learning is an emerging area.
Providing visual context has been explored by~\cite{bar}, Painter~\cite{painter}, SegGPT~\cite{seggpt} and PromptGIP~\cite{promptgip}. They propose elegant solutions that involve unifying the output space of a network to solve multiple computer vision tasks based on visual context. The core idea of these frameworks is the usage of masked image modelling (MIM) to extract context from the unmasked regions of an image and subsequently employing in-painting for the desired prediction. Although Painter was trained for image restoration tasks using visual in-context learning, we show that it fails to use contextual information effectively, hindering its restoration performance. We believe this is due to two primary reasons. Firstly, the extraction of context relies solely on the MIM framework, which lacks constraints to ensure retrieval of degradation characteristics from the context images. Secondly, providing context only at the encoder's input can suppress contextual information after the initial network layers, making it negligible at the decoder. 
PromptGIP also suffers from the above issues, leading to inferior performance.

We propose AWRaCLe, a methodology for all-weather (rain, snow and haze) image restoration which elegantly leverages visual in-context learning. Our method aims to restore a query image by utilizing additional context (that we call context pair), which comprises of a degraded image and its corresponding clean version (see Fig.~\ref{fig:first}). The context requires paired images since without scene consistency, it is challenging to extract DSI which will in-turn hamper restoration performance. To facilitate restoration, the degradation type in the context pair should align with that of the query image. During test time, the context pair for a degradation can be chosen from the respective training set, thus requiring only the knowledge of the degradation type. 

We devise Degradation Context Extraction (DCE) blocks that leverage features from CLIP's~\cite{clip} image encoder and employ self-attention mechanisms to extract relevant DSI, such as the type and visual characteristics of the degradation, from the given context pair. Additionally, we introduce Context Fusion (CF) blocks designed to integrate the extracted context from the DCE blocks with the feature maps of an image restoration network. Specifically, we use the Restormer~\cite{restormer} network for this purpose. The fusion process involves multi-head cross attention at each spatial level of the decoder, ensuring the propagation of context information throughout the restoration network, thus enhancing the performance. Representative results on haze, snow and rain removal are given in Fig.~\ref{fig:first} to demonstrate the efficacy of our method. Interestingly, AWRaCLe can harness DSI from the context pair to perform selective degradation removal. For instance in Fig.~\ref{fig:first}d and e, a query image corrupted by both haze and snow is presented as input. In Fig.~\ref{fig:first}d, the context pair is given as haze and AWRaCLe returned an output that contained only snow. Similarly, in Fig.~\ref{fig:first}e, the context pair is given as snow resulting in an output that contained only haze. This reaffirms the ability of AWRaCLe in utilising degradation-context effectively. We have performed extensive experiments to demonstrate the effectiveness of AWRaCLe and show that our method achieves state-of-the-art performance for the AWIR task. 


In summary, our main contributions are as follows:
\begin{enumerate}
    \item  We propose a novel approach called AWRaCLe that employs visual in-context learning for AWIR. To the best of our knowledge, this is the first work which effectively utilizes visual degradation context for AWIR.
    \item  We propose novel Degradation Context Extraction blocks and Context Fusion blocks which extract and fuse relevant degradation information from the provided visual context. Our method ensures that the extracted context is injected suitably at different stages of the restoration network to enable context information flow.
    \item Through comprehensive experiments, we show that AWRaCLe achieves state-of-the-art all-weather restoration performance on multiple benchmark datasets.
\end{enumerate}

\section{Related Works}
\label{sec:related_works}

In this section, we discuss relevant works on adverse weather restoration and in-context learning.

\subsection{Adverse weather restoration}
Several methods have been proposed for single weather restoration such as ~\cite{rain1, rain2} for deraining,~\cite{haze1, haze2} for dehazing and ~\cite{snow1, snow2} for desnowing. Recently, methods such as Restormer~\cite{restormer} and MPRNet~\cite{mprnet} have tackled multiple degradations. However, the above methods require retraining or fine-tuning for each degradation. To overcome this limitation, AWIR methods have been actively explored. All-in-one~\cite{nas} used neural architecture search to find the best-suited encoder for each degradation from a set of encoders. Transweather~\cite{transw} employed a unified network with a single encoder for multi-weather restoration. Airnet~\cite{airnet} used Momentum Contrast (MoCo)~\cite{moco} for improved degradation representations while TSMC~\cite{tkmc} proposed two-stage knowledge learning with multi-contrastive regularization for a similar objective. Recently, WeatherDiff~\cite{weatherdiff} proposed a patch-based denoising diffusion model for adverse weather removal and WGWS~\cite{wgws} extracted weather-general and weather-specific features for restoration. PromptIR~\cite{promptir} utilized learnable prompt embeddings for AWIR while DiffUIR~\cite{diffuir} proposed selective hourglass mapping. These AWIR methods do not utilize contextual guidance which limits their performance. Diff-Plugin~\cite{diffplugin} and DA-CLIP~\cite{daclip} attempt to extract degradation-specific information (DSI) directly from images to improve performance. However, this approach makes it challenging to disentangle scene characteristics from DSI, unlike our method.

\begin{figure*}[t]
    \centering
        \includegraphics[height=8cm, width=15.5cm]{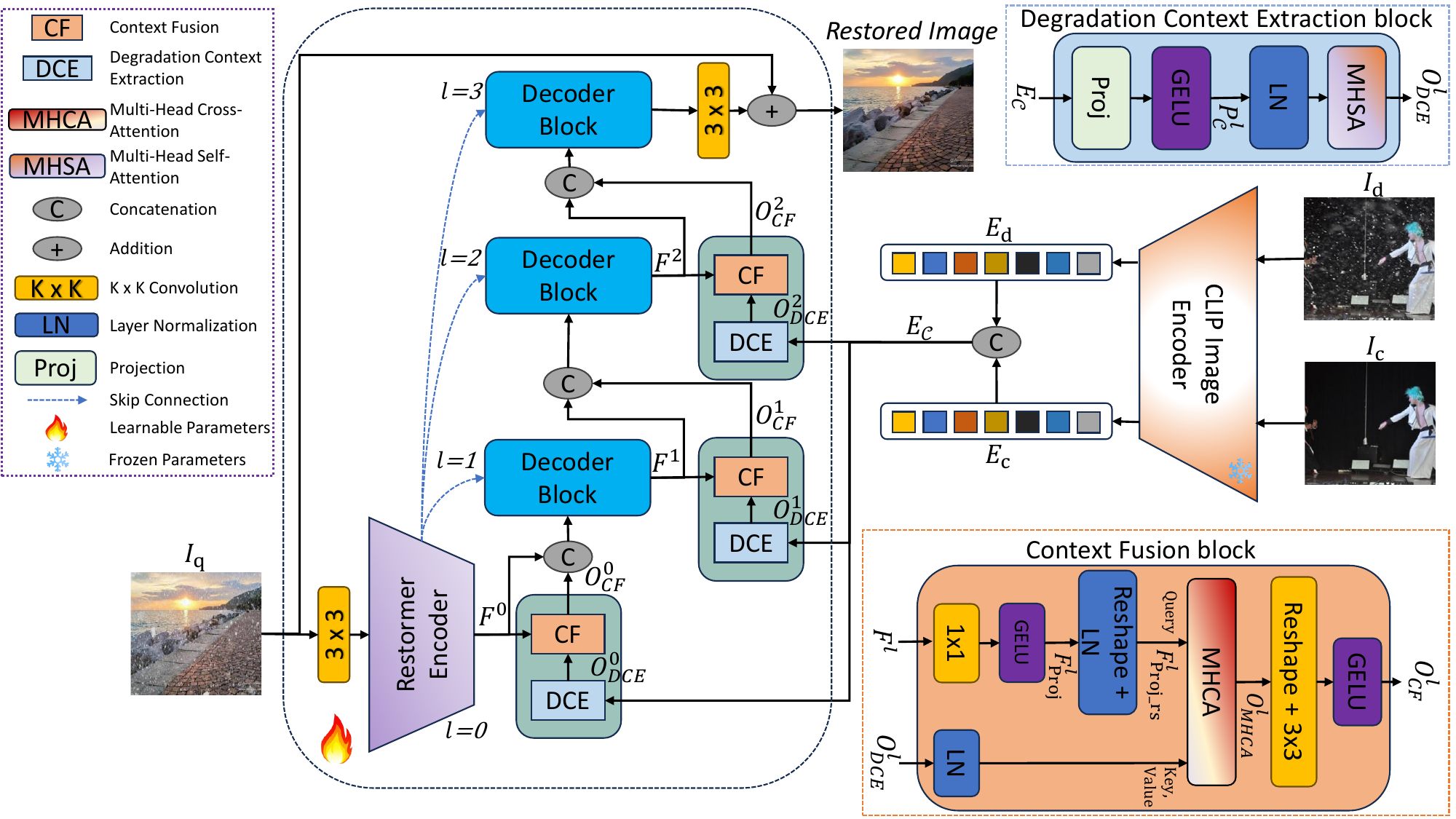}
   \caption{Block diagram of the proposed visual in-context learning approach for AWIR. CLIP features are extracted from $I_{\texttt{d}}$ and $I_{\texttt{c}}$ which are subsequently fed to DCE blocks at different decoder levels, $l$. CF blocks then fuse the degradation information obtained from the DCE blocks with decoder features, $F^{l}$, from the query image $I_{\texttt{q}}$. Finally, the restored image is generated.}
    \label{fig:outline}
\end{figure*}

\subsection{In-context Learning}
Transformers have a generalized modeling capability through the use of tokens. Leveraging this, DETR~\cite{detr} used transformer heads for object detection. Pix2Seq~\cite{pix2seq} discretized the output space of object detection. Unified-IO~\cite{uio} and Pix2Seqv2~\cite{pix2seqv2} extended this approach to multiple vision tasks (generalist models) using task prompts. These methods use a discrete output space, which is unsuitable for continuous space of image data, making it challenging to enable visual in-context learning. 

Recent advancements in in-context learning have significantly improved the zero-shot performance of large language models. GPT-3~\cite{gpt3} demonstrated this using text completion with prompts as context while Flamingo~\cite{flamingo} used language guidance for various image and video tasks. Visual in-context learning is an emerging area which is gaining increasing attention. VPI~\cite{bar} proposed an image-based continuous output space for visual in-context segmentation. Directly using visual context for tackling multiple computer vision tasks is challenging due to the non-unified output space. To address this, Painter~\cite{painter} extended VPI for diverse computer vision tasks by unifying their output spaces. PromptGIP~\cite{promptgip} proposed a visual prompting question-answering framework for extracting context. However, these approaches rely heavily on masked-image modelling to learn context (see Painter), which is ineffective for image restoration because there is no dedicated module to capture degradation-specific information. Additionally, context information provided at the network's input may not propagate to deeper layers.


\section{Proposed Methodology}
\label{sec:proposed_method}
In this section, we explain in detail our proposed approach, AWRaCLe, for performing AWIR (deraining, desnowing and dehazing) using visual in-context learning. A high-level schematic of AWRaCLe is shown in Fig.~\ref{fig:outline}. The main idea of our approach involves extracting relevant degradation-context such as the type and visual characteristics of degradations from a given image-ground truth pair to effectively restore a query image with the same type of degradation. Toward this aim, we propose Degradation Context Extraction (DCE) and Context Fusion (CF) blocks that learn context information and fuse it with an image restoration network to facilitate the restoration process. Specifically, we integrate our DCE and CF blocks with a slightly modified version of the Restormer network (see supplementary for details). The DCE and CF blocks are added at each decoder level of Restormer for propagation of context information (multi-level fusion). The decoder levels are represented by $l=0, 1, 2$ and $3$ in Fig.~\ref{fig:outline}. AWRaCLe overcomes the limitations of Painter which solely relies on masked image modelling to extract context information. Also, they provide context information only at the input, thus lacking any mechanism to ensure its flow throughout the network. 

\noindent \textbf{Terminology.} For ease of understanding, we define a few terms. We refer to the context-pair as $\mathcal{C}=\{I_{\texttt{d}}, I_{\texttt{c}}\}$ where $I_{\texttt{d}}$ is the degraded image and $I_{\texttt{c}}$ is its corresponding clean image. $I_{\texttt{q}}$ is the degraded query image which needs to be restored given $\mathcal{C}$. Note that $I_{\texttt{q}}$ and $I_{\texttt{d}}$ are affected by the same type of degradation. Additionally, $I_{\texttt{d}}, I_{\texttt{c}}, I_{\texttt{q}} \in \mathbb{R}^{H\times W\times 3}$ where $H, W$ indicate the spatial resolution of the images. 

\subsection{Degradation Context Extraction}
\label{subsec:dce}
The objective of the DCE blocks is to extract degradation-specific context such as the type and visual characteristics from $\mathcal{C}$. It is crucial for the underlying scene content in $I_{\texttt{d}}$ and $I_{\texttt{c}}$ to be identical so that the only distinction between them is the degradation (we call this paired context). This condition facilitates the process of extracting degradation-specific information (DSI) from $I_{\texttt{d}}$ and $I_{\texttt{c}}$. In the ablations, we show that using un-paired context ($I_{\texttt{d}}$ and $I_{\texttt{c}}$ are from different scenes) leads to inferior performance. Furthermore, it is important to note the considerable difficulty in extracting degradation-context solely from $I_{\texttt{d}}$. This challenge arises since it is not straightforward to disentangle the scene content from the degradation. 

We now elaborate the aforementioned process of extracting degradation-context from $\mathcal{C}$. Vision-Language models (VLMs) such as CLIP~\cite{clip} have demonstrated the capability to learn high quality image embeddings that can be used to solve a myriad of downstream computer vision tasks~\cite{clip_appl1,clip_appl2,clip_appl3,clip_appl4}. 
Motivated by this, we obtain representations $E_{\texttt{d}}, E_{\texttt{c}}$  for the context pair $\mathcal{C}$ from the final transformer block of CLIP's image encoder (denoted by $\texttt{CLIP} (.)$). The obtained features are then fed to the DCE blocks that are present at each decoder level $l$ of the network. The above feature extraction step using CLIP is given as follows.
\small
\begin{equation}
\label{eq:clip}
\begin{split}
E_{\texttt{d}}=\texttt{CLIP}(I_{\texttt{d}}), E_{\texttt{c}} = \texttt{CLIP}(I_{\texttt{c}}), \{E_{\texttt{d}}, E_{\texttt{c}}\} \in \mathbb{R}^{L\times D}
\end{split}
\end{equation}
\normalsize
$E_{\texttt{d}}$ and $E_{\texttt{c}}$ are then concatenated to obtain $E_{\mathcal{C}} \in \mathbb{R}^{2L\times D}$ as the overall CLIP representation for $\mathcal{C}$. Here, $L$ represents the number of tokens and $D$ is the embedding dimension. Within a DCE block at level $l$, $E_{\mathcal{C}}$ is initially projected to a lower dimension to reduce computational complexity for forthcoming attention operations. This is followed by GELU~\cite{gelu} activation function as non-linearity. The result of these operations is $P_{\mathcal{C}}^{l} \in \mathbb{R}^{2L\times C^{l}}$, where $C^l$ is the projection dimension, and these steps are summarised as below.
\small
\begin{equation}
\label{eq:clip_proj}
    P_{\mathcal{C}}^{l} = \texttt{GELU}(\texttt{Proj}(E_{\mathcal{C}})), P_{\mathcal{C}}^{l} \in \mathbb{R}^{2L\times C^{l}}
\end{equation}
\normalsize
The projected feature, $P_{\mathcal{C}}^{l}$, is normalized using layer normalization ($\texttt{LN}$)~\cite{ln}. Subsequently, Multi-Head Self-Attention ($\texttt{MHSA}(.)$)~\cite{vit_sa} is employed to capture DSI, $O_{\texttt{DCE}}^{l}$, from $P_{\mathcal{C}}^{l}$. This step can be summarized  as
\begin{equation}
    \label{eq:clip_self}
    O_{\texttt{DCE}}^{l} = \texttt{MHSA}(\texttt{LN}(P_{\mathcal{C}}^{l})), O_{\texttt{DCE}}^{l} \in \mathbb{R}^{2L\times C^{l}}.
\end{equation}
Since the scene is consistent in both $I_{\texttt{d}}$ and $I_{\texttt{c}}$, the primary distinction between them is the degradation which is adeptly discerned through MHSA. This enables extraction of the necessary degradation-context from ${\mathcal{C}}$ for judiciously guiding the network towards the objective of all-weather restoration. Fig.~\ref{fig:outline} shows a detailed schematic of the DCE block, highlighting the above steps. Additional details about the MHSA module are given in the supplementary document.

\begin{figure}[t]
    \centering
    \begin{subfigure}[h]{1\linewidth}
        \centering
        \small
        \setlength{\tabcolsep}{1pt}
        \begin{tabular}{ccccc}
             &$I_\texttt{d}$&$I_\texttt{c}$&$I_\texttt{d}+A_\texttt{d}$&$I_\texttt{c}+A_\texttt{c}$\\
             \rotatebox[origin=tl]{90}{Haze}&\includegraphics[height=1.1cm, width=0.226\linewidth]{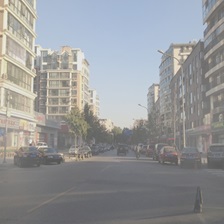}&\includegraphics[height=1.1cm, width=0.226\linewidth]{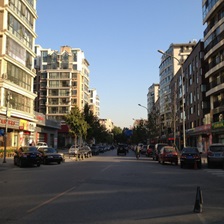}&\includegraphics[height=1.1cm, width=0.226\linewidth]{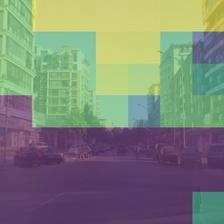}&          \includegraphics[height=1.1cm, width=0.226\linewidth]{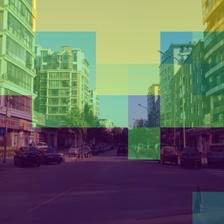} \\

             \rotatebox[origin=tl]{90}{Snow}&
             \includegraphics[height=1.1cm, width=0.226\linewidth]{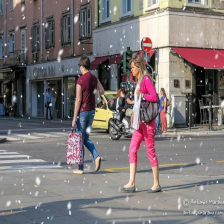}&\includegraphics[height=1.1cm, width=0.226\linewidth]{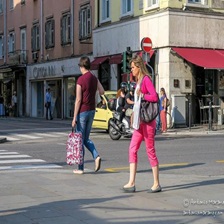}&\includegraphics[height=1.1cm, width=0.226\linewidth]{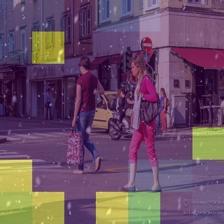}&          \includegraphics[height=1.1cm, width=0.226\linewidth]{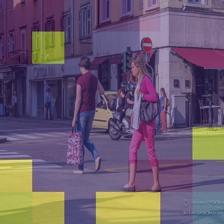}
        \end{tabular}
       \caption{DCE block activations $A_\texttt{d}$ and $A_\texttt{c}$ overlayed ($+$) on $I_\texttt{d}$ and $I_\texttt{c}$, respectively, of the context pair. Yellow-High, Blue-Low}
        \label{fig:DCE_overlay}
    \end{subfigure}

    \begin{subfigure}[t]{1\linewidth}
        \fbox{\includegraphics[height=1.2cm, width=0.45\linewidth]{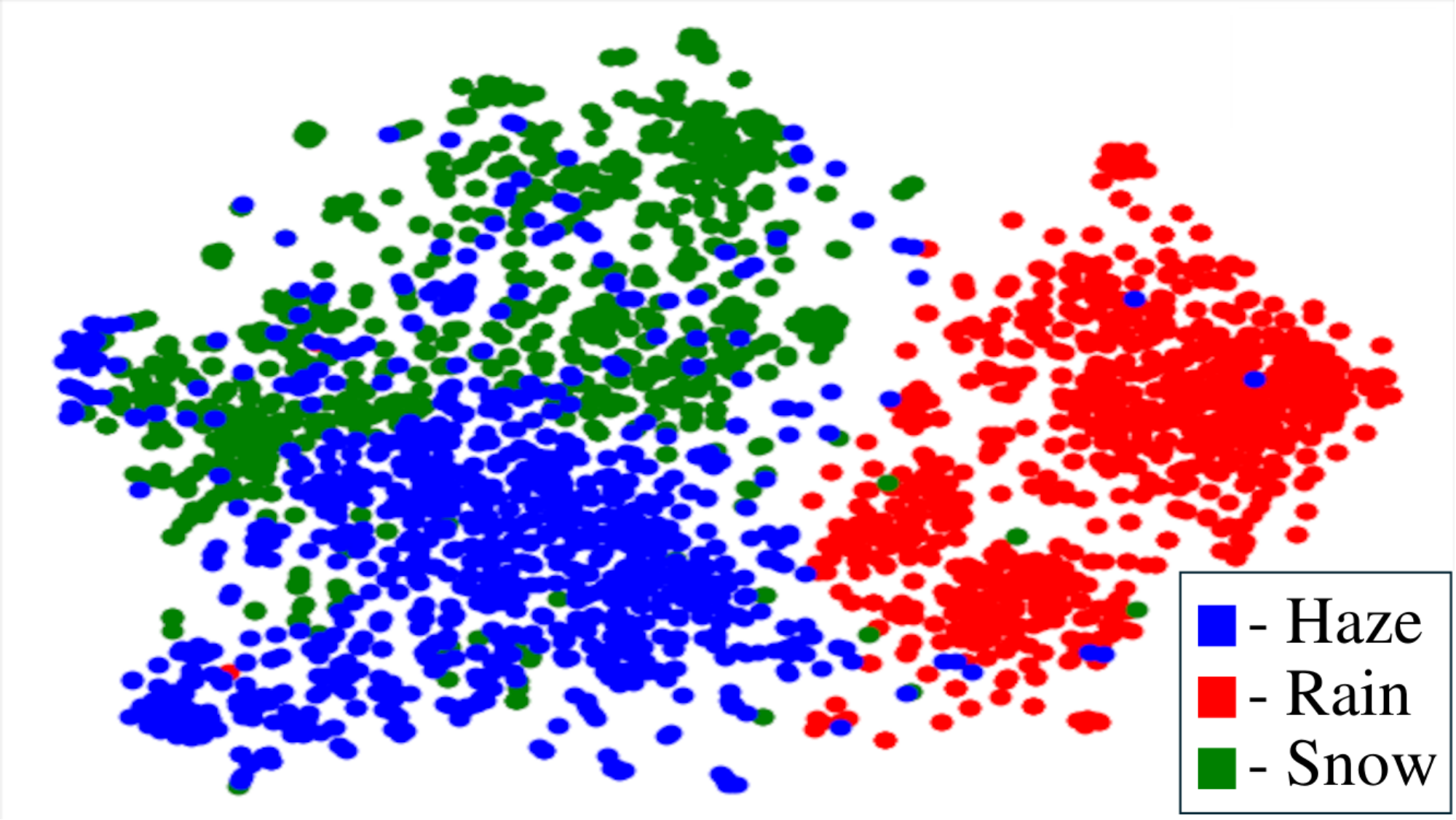}}
         \fbox{\includegraphics[height=1.2cm, width=0.45\linewidth]{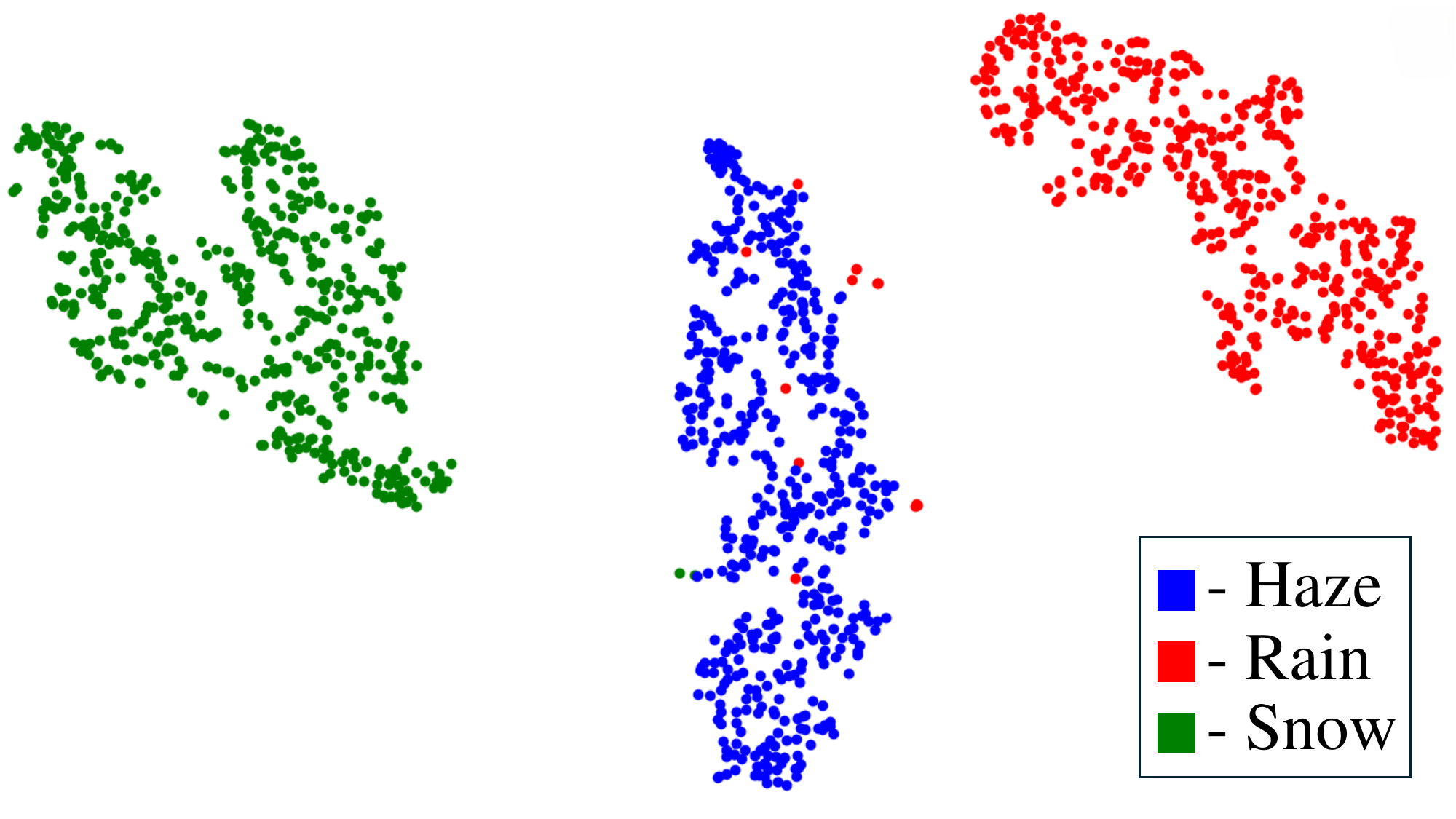}}
        \caption{t-SNE plot of CLIP embeddings ($E_\mathcal{C}$, right) and DCE block outputs ($O_{\texttt{DCE}}^{l}$, left), for hazy, rainy and snowy context pairs. Separation signficantly improves after using the DCE block.}
    \label{fig:DCE_tsne}
    \end{subfigure}
    \caption{Analysis of DCE block outputs.}
\end{figure}

To visualize the extracted degradation-specifc information, we overlay the output of the DCE block ($O_{\texttt{DCE}}^{l}$) for a clean-hazy and a clean-snowy context pair, respectively. This is illustrated in Fig.~\ref{fig:DCE_overlay} where activations are overlayed on the degraded image ($I_\texttt{d}$) and its corresponding clean image ($I_\texttt{c}$). $A_\texttt{d} \in \mathbb{R}^{L\times C^l}$ and $A_\texttt{c} \in \mathbb{R}^{L\times C^l}$ represent the DCE block activations obtained by splitting $O_{\texttt{DCE}}^{l}$ for $I_\texttt{d}$ and $I_\texttt{c}$, respectively. The figure shows that the DCE block captures DSI such as the spatially-varying characteristics of haze and sparseness of snow. Furthermore, to discern this information, the DCE block uses the clean image ($I_\texttt{c}$) to identify and focus on degraded regions in $I_\texttt{d}$, evident from attention at similar locations in both $I_\texttt{d}$ and $I_\texttt{c}$. Additionally, Fig.~\ref{fig:DCE_tsne} provides t-SNE plots of the CLIP embeddings, $E_\mathcal{C}$, and DCE block outputs ($O_{\texttt{DCE}}^{l}$) for hazy, snowy and rainy context pairs. Although there is separation in the t-SNE plot with $E_\mathcal{C}$, it is significantly enhanced after using the DCE block. Thus, the DCE block extracts DSI that is clustered closely for the same type of degradation but is separated for different degradations.

\subsection{Context Fusion}
\label{subsec:cf}

At each level $l$, the obtained degradation-context, $O_{\texttt{DCE}}^{l}$, needs to be fused with the corresponding decoder features, $F^{l} \in \mathbb{R}^{K^{l} \times H^{l} \times W^{l}}$, from Restormer. Here $H^{l}$ and $W^{l}$ are the spatial resolution of the feature map, and $K^{l}$ is channel dimension. Fusion is achieved with the help of the Context Fusion (CF) blocks that utilize Multi-Head Cross Attention ($\texttt{MHCA(.)}$)~\cite{vit_ca} to integrate information from $O_{\texttt{DCE}}^{l}$ and $F^{l}$. The cross-attention mechanism is a key ingredient in the CF module as we want $F^{l}$ to be enhanced by the degradation information contained in $O_{\texttt{DCE}}^{l}$. We achieve this by treating $F^{l}$ as the query and matching it with the key and value computed from $O_{\texttt{DCE}}^{l}$. 

The CF module which is illustrated in Fig.~\ref{fig:outline} is next described in detail. Prior to MHCA, $F^{l}$ is projected to the same channel dimension ($C^{l}$) as $O_{\texttt{DCE}}^{l}$ using $1\times 1$ convolution as 
\small
\begin{equation}
\label{eq:cf_conv1x1}
    F^{l}_\texttt{Proj} = \texttt{GELU}(\texttt{Conv}_{1\times 1}(F^{l})), F^{l}_\texttt{Proj} \in \mathbb{R}^{C^{l} \times H^{l} \times W^{l}},
\end{equation}
\normalsize
where we have used the $\texttt{GELU}$ activation function as non-linearity.  We observe that $F^{l}_\texttt{Proj}$ and $O_{\texttt{DCE}}^{l}$ have a mismatch in the number of  dimensions ($O_{\texttt{DCE}}^{l}$ is $2$-D but $F^{l}_\texttt{Proj}$ is $3$-D), which precludes the use of standard MHCA operation. One plausible approach involves the use of reshaping operations followed by interpolation to transform $O_{\texttt{DCE}}^{l}$ into the same dimension as $F^{l}_\texttt{Proj}$. However, interpolation causes redundancy in the degradation-context thereby hindering the performance of cross-attention. To circumvent this problem, we reshape (denoted as $\texttt{RH(.)}$)
$F^{l}_\texttt{Proj}$ to obtain $F^{l}_\texttt{Proj\_rs} \in \mathbb{R}^{H^{l}\cdot W^{l} \times C^{l}}$ which has the same channel dimensions, $C^{l}$, as $O_{\texttt{DCE}}^{l}$. Notice that no interpolation operations are required as the number of dimensions are now consistent between $O_{\texttt{DCE}}^{l}$ and $F^{l}_\texttt{Proj\_rs}$. Subsequently, layer normalization is applied to both $O_{\texttt{DCE}}^{l}$ and $F^{l}_\texttt{Proj\_rs}$. The above steps can be summarised as
\begin{figure}[t]
    \centering
    \setlength{\tabcolsep}{1pt}
    \small
    \begin{tabular}{ccc}
         $I_\texttt{q}$& Before CF ($F^{l}_\texttt{Proj}$)&After CF ($O_{\texttt{CF}}^{l}$) \\
         \includegraphics[height=1.19cm, width=0.27\linewidth]{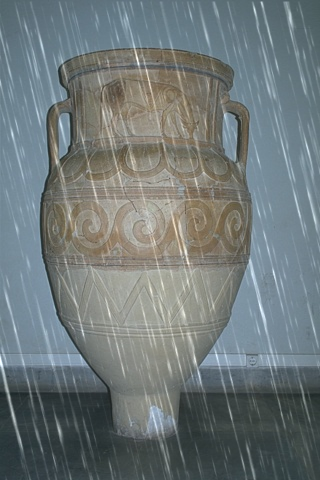}& 
         \includegraphics[height=1.19cm, width=0.27\linewidth]{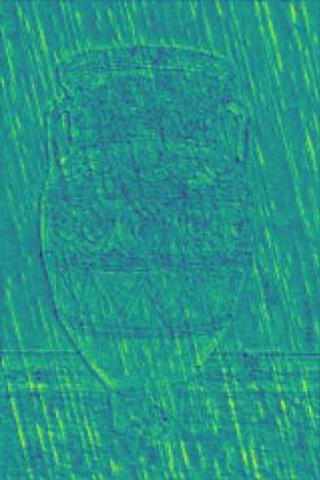}& \includegraphics[height=1.19cm, width=0.27\linewidth]{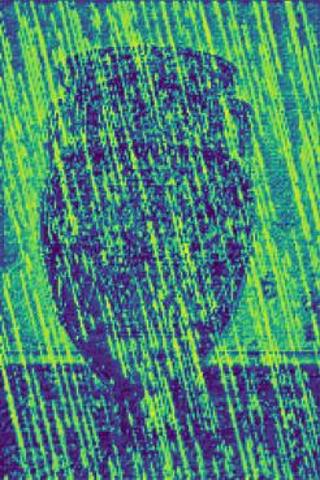}
    \end{tabular}
   \caption{Comparison of activations of the restoration network prior to CF and after CF. Yellow-High, Blue-Low.}
    \label{fig:CF_paired}
\end{figure}

\small
\begin{equation}
    \label{eq:mhca_ln}
    F^{l}_\texttt{Proj\_rs} = \texttt{LN}(\texttt{RH}(F^{l}_\texttt{Proj})), O_{\texttt{DCE}}^{l} = \texttt{LN}(O_{\texttt{DCE}}^{l}).
\end{equation}
\normalsize
Next, we employ cross-attention to integrate relevant degradation-specific
information into $F^{l}_\texttt{Proj\_rs}$ and this is achieved using MHCA. For this purpose, we use $F^{l}_\texttt{Proj\_rs}$ to calculate the query ($\texttt{Q}$) and $O_{\texttt{DCE}}^{l}$ for computing the key ($\texttt{K}$) and value ($\texttt{V}$) as follows
\small
\begin{equation}
        \label{eq:mhca_qkv}
        F^{l}_\texttt{Proj\_rs} \xrightarrow{} \texttt{Q}, O_{\texttt{DCE}}^{l} \xrightarrow{} \texttt{K}, \texttt{V},
    \end{equation}
    \begin{equation}
        \label{eq:mhca_fuse}
        O_{\texttt{MHCA}}^{l} = \texttt{MHCA}(\texttt{Q}, \texttt{K}, \texttt{V}), O_{\texttt{MHCA}}^{l} \in \mathbb{R}^{H^{l}\cdot W^{l} \times C^{l}}.
\end{equation}
\normalsize
\begin{table*}[t]
\small
\centering
\caption{Quantitative comparisons of AWRaCLe with SOTA on the test sets described in Sec.~\ref{subsec:datasets}. The values indicated are placeholders for PSNR/SSIM. The best result is in \textbf{bold}, and second best is \underline{underlined}.}

\setlength{\tabcolsep}{1.2pt}
\begin{tabularx}{\textwidth}{l *{11}{c}}
\toprule
Datasets & WeatherDiff & WGWS & TSMC & AirNet & PromptIR & Painter & DA-CLIP & DiffUIR & DiffPlugin & PromptGIP & AWRaCLe \\
& TPAMI'23    & CVPR'23 & CVPR'22 & CVPR'22 & NeurIPS'23 & CVPR'23 & ICLR'24  & CVPR'24   & CVPR'24    & ICML'24 & - \\

\midrule
SOTS & 28.0/0.966 & 30.5/0.976 & 27.9/0.920 & 27.6/0.963 & 30.5/0.977 & 28.0/0.945 & 26.9/0.958 & \underline{31.0}/\underline{0.977} & 23.6/0.778 & 17.9/0.672 & \textbf{31.7}/\textbf{0.981} \\

Rain100H & 25.8/\underline{0.824} & 13.9/0.410 & 26.5/0.822 & 23.0/0.692 & 26.3/0.821 & 22.5/0.792 & 23.4/0.730 & \underline{26.5}/0.788 & 16.1/0.527 & 18.0/0.482 & \textbf{27.2}/\textbf{0.840} \\

Rain100L & 27.4/0.895 & 27.2/0.860 & 29.9/0.920 & 24.0/0.805 & 28.9/0.888 & 23.2/0.900 & 30.1/0.918 & \underline{31.8}/\underline{0.932} & 25.4/0.698 & 22.8/0.662 & \textbf{35.7}/\textbf{0.966} \\

Snow100k & 31.3/0.910 & 32.6/0.921 & 32.3/0.916 & 29.2/0.884 & \underline{33.4}/\underline{0.932} & 27.9/0.871 & 30.6/0.893 & 31.8/0.915 & 23.5/0.658 & 20.8/0.615 & \textbf{33.5}/\textbf{0.934} \\

\midrule
Average & 28.1/0.898 & 26.0/0.791 & 29.1/0.894 & 25.9/0.836 & 29.8/\underline{0.904} & 26.4/0.877 & 27.8/0.874 & \underline{30.3}/0.903 & 22.2/0.665 & 19.9/0.608 & \textbf{32.0}/\textbf{0.930} \\
\bottomrule
\end{tabularx}
\label{tab:quant_train}
\end{table*}
We select $F^{l}_\texttt{Proj\_rs}$ as the query since we are looking to match the relevant DSI from $O_{\texttt{DCE}}^{l}$ (key) to enhance the feature maps with the extracted context information. 

The output of MHCA, is then reshaped back to $\mathbb{R}^{C^{l} \times H^{l} \times W^{l}}$ and is projected using $3\times 3$ convolution ($\texttt{Conv}_{3\times 3}$). Again, GELU activation is applied to obtain the output of the CF block, $O_{\texttt{CF}}^{l}$ (see Eqn.~\ref{eq:mhca_out}). More details about the workings of MHCA are provided in the supplementary document.
\small
\begin{equation}
    \label{eq:mhca_out}
        O_{\texttt{CF}}^{l} = \texttt{GELU}(\texttt{Conv}_{3\times 3}(\texttt{RH}(O_{\texttt{MHCA}}^{l}))), O_{\texttt{CF}}^{l} \in \mathbb{R}^{C^{l}\times H^{l}\times W^{l}}.
\end{equation}
\normalsize
Finally, $O_{\texttt{CF}}^{l}$ is concatenated with $F^{l}$ and propagated to the next decoder level. Fig.~\ref{fig:CF_paired} captures the activations from the network for a rainy image ($I_\texttt{q}$) prior to CF ($F^{l}_\texttt{Proj}$) and after CF ($O_{\texttt{CF}}^{l}$). Observe that prior to CF, not much attention is paid to degraded regions (rain streaks). However, after CF, the attention increases significantly on the rain streaks of $I_\texttt{q}$. Thus, the CF block effectively fuses the DSI ($O_{\texttt{DCE}}^{l}$) into the features of the restoration network ($F^l$).

The process of degradation-context extraction and context fusion is repeated at each level, $l$, of the decoder. This multi-scale fusion at each decoder level $l$, ensures that the context information is retained through the entire decoder, thereby enhancing the quality of image reconstruction.

\section{Experimental Results}
\label{sec:experiments}
In this section, we explain our implementation, datasets used, results and ablation studies.
\subsection{Implementation Details}
Our method is trained using the AdamW optimizer with a cosine annealing Learning Rate (LR) scheduler. We train for a total of $100$ epochs on $8$ RTX A5000 GPUs with a batch size of $32$, initial LR$=2\times 10^{-4}$, weight decay$=0.01$, $\beta_{1}=0.9$, $\beta_{2}=0.999$ and warm-up for $15$ epochs. We use random crop size of $128\times 128$ pixels, and random flipping as data augmentations. The loss function used is the $L_{1}$ loss. For extracting CLIP features, no augmentations are used and the images in the context pair are resized to $224\times224$. Our implementation utilized PyTorch~\cite{pytorch}.

\subsection{Datasets}
\label{subsec:datasets}
\textbf{Training.} We use the Snow100k~\cite{snow100k}, synthetic rain~\cite{mprnet} datasets (SRD) and RESIDE~\cite{reside} to train our method for all-weather restoration. The training split of Snow100k contains $50,000$ synthetic snow images along with the corresponding clean images. For deraining, we use the training split of SRD containing $13,711$ clean-synthetic rainy image pairs. For dehazing, we use the Outdoor Training Set (OTS) of RESIDE which consists of $72,135$ clean-synthetic hazy image pairs for training. We then split the training sets into two categories, each respectively consisting of heavy and light corruptions for better context extraction during training. More details about the splitting strategy can be found in the supplementary. In summary, we obtain $12,077$ light rain images, $1,634$ heavy rain images, $38,921$ light haze images, $33,214$ heavy haze images, $37,122$ light snow images and $12,878$ heavy snow images for training.\\

\noindent \textbf{Evaluation. }We evaluate all the methods for desnowing, deraining and dehazing. For desnowing, we use the test split of Snow100k dataset containing $50,000$ paired images. For deraining, we evaluate the methods on Rain100H~\cite{rain100handl} for heavy rain and Rain100L~\cite{rain100handl} for light rain, each consisting of $100$ paired images. For dehazing, we use RESIDE's Synthetic Objective Testing Set (SOTS) outdoor containing $500$ paired images.

\subsection{Comparisons}
\label{subsec:comp_disc}
We evaluate and compare the performance of AWRaCLe with ten recent AWIR approaches on the test sets described in Sec.~\ref{subsec:datasets}. The methods we use for comparison are WeatherDiff~\cite{weatherdiff}, WGWS~\cite{wgws}, TSMC~\cite{tkmc}, AirNet~\cite{airnet}, PromptIR~\cite{promptir}, DiffUIR~\cite{diffuir}, DiffPlugin~\cite{diffplugin} and DA-CLIP~\cite{daclip}. Additionally, we also compare with Painter~\cite{painter} and PromptGIP~\cite{promptgip}, and show that AWRaCLe uses context much more effectively. For a fair comparison, all methods are retrained on the training sets mentioned in Sec.~\ref{subsec:datasets}. Some recent approaches such as~\cite{domtrans,distillsam,ptpriors,mpperceiver} have no training code. Hence, we are unable to compare with these methods. We also do not compare with methods for single weather removal as our method is proposed specifically to deal with multiple degradations.


\noindent \textbf{Quantitative and Qualitative results.}
We discuss the performance of all the methods on the test sets described in Sec.~\ref{subsec:datasets}. Table~\ref{tab:quant_train} contains the Peak Signal to Noise Ratio (PSNR) and Structural Similarity (SSIM) values for each method on these test sets. To evaluate AWRaCLe, Painter and PromptGIP, we choose the context pair for each test set randomly from their respective training sets, i.e., for every test image, the context pair is chosen randomly. We resort to random selection for fairness. 
Moreover, the context pair is chosen from the training set, thus, requiring only the knowledge of the type of degradation during inference. From Table~\ref{tab:quant_train}, we observe that AWRaCLe achieves excellent overall metrics. Our approach yields highest PSNR and SSIM values across all datasets. Importantly, AWRaCLe offers consistently high performance across all restoration tasks whereas competing methods perform well for some tasks but poorly for others. We also significantly outperform the in-context learning approaches, Painter and PromptGIP, highlighting the effectiveness of our in-context learning strategy. Additionally, we provide quantitative comparisons with LPIPS and FID scores in the supplementary. 

In Fig.~\ref{fig:qual}, we show qualitative results for visual inspection and compare with the top-performing approaches TSMC, PromptIR and DiffUIR. It can be observed  that AWRaCLe is able to handle the corruptions more effectively than the others. More qualitative results, performance of AWRaCLE on real images along with a user study, and a discussion of the limitations of our method are provided in the supplementary. 

\begin{figure*}[t]
\small
\setlength\tabcolsep{1pt}
    \centering
    \begin{tabular}{ccccccc}
    
        &Input&TSMC&PromptIR&DiffUIR&AWRaCLe&GT\\


        \rotatebox[origin=c]{90}{SOTS\hspace{-40pt}}&\includegraphics[height=0.09\linewidth, width=0.15\linewidth]{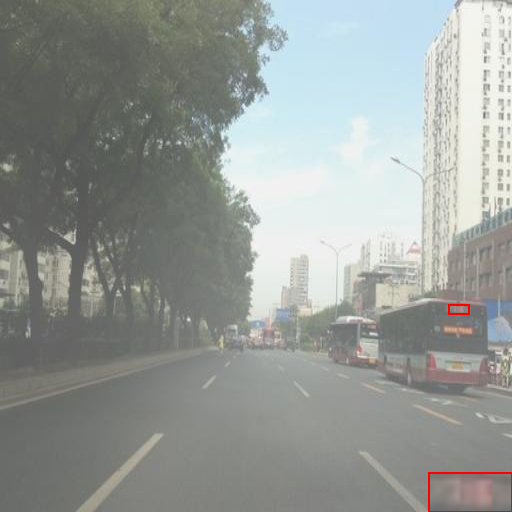}&\includegraphics[height=0.09\linewidth, width=0.15\linewidth]{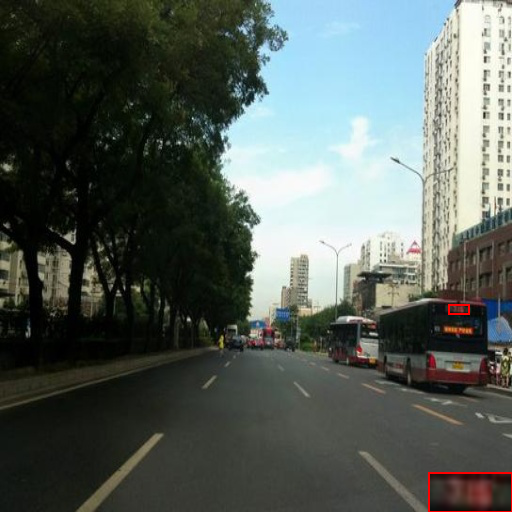}&\includegraphics[height=0.09\linewidth, width=0.15\linewidth]{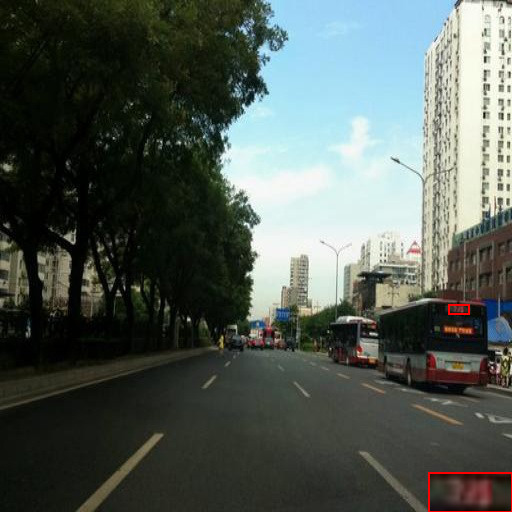}&\includegraphics[height=0.09\linewidth, width=0.15\linewidth]{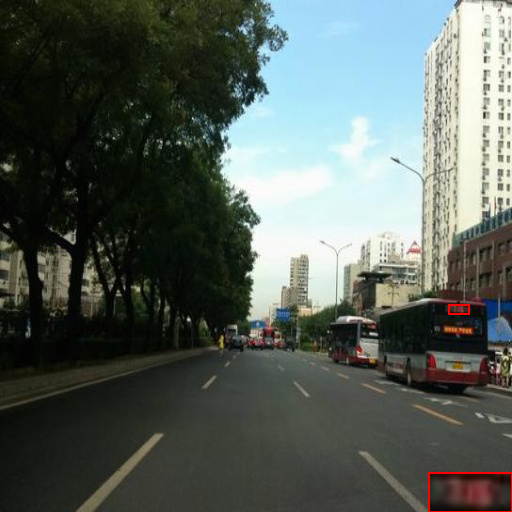}&\includegraphics[height=0.09\linewidth, width=0.15\linewidth]{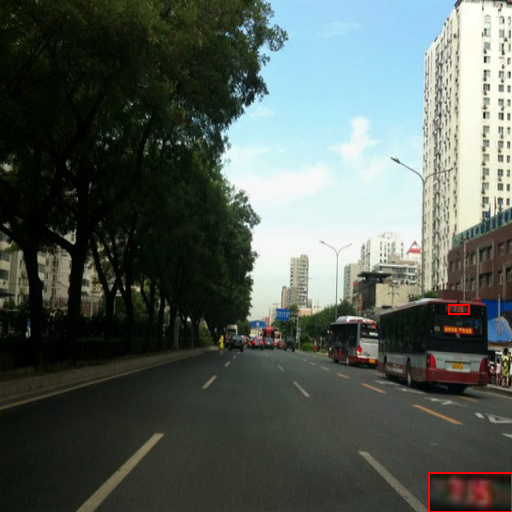}&\includegraphics[height=0.09\linewidth, width=0.15\linewidth]{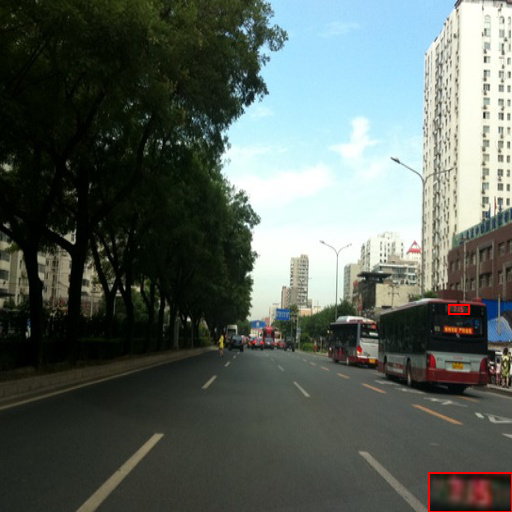}\\


        \rotatebox[origin=c]{90}{Rain100H\hspace{-40pt}}&\includegraphics[height=0.09\linewidth, width=0.15\linewidth]{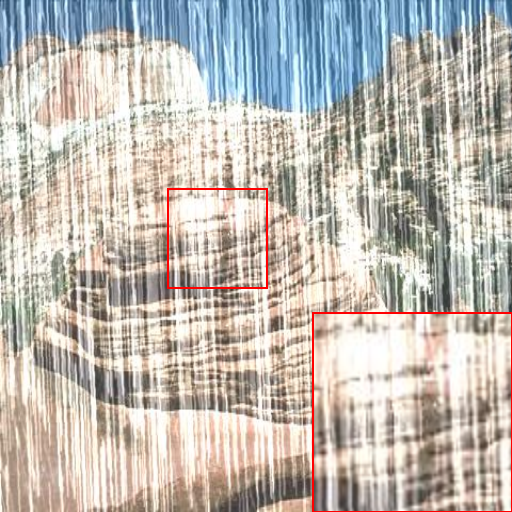}&\includegraphics[height=0.09\linewidth, width=0.15\linewidth]{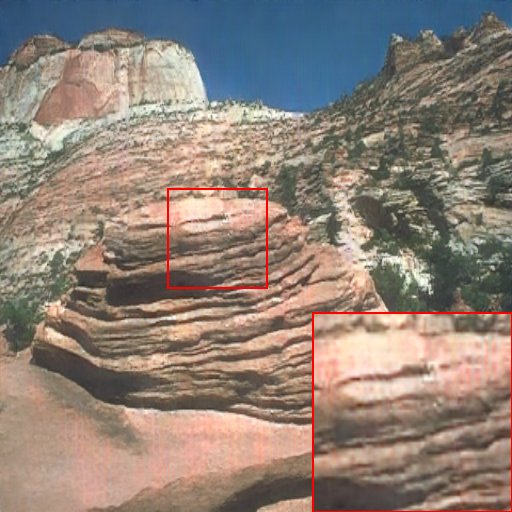}&\includegraphics[height=0.09\linewidth, width=0.15\linewidth]{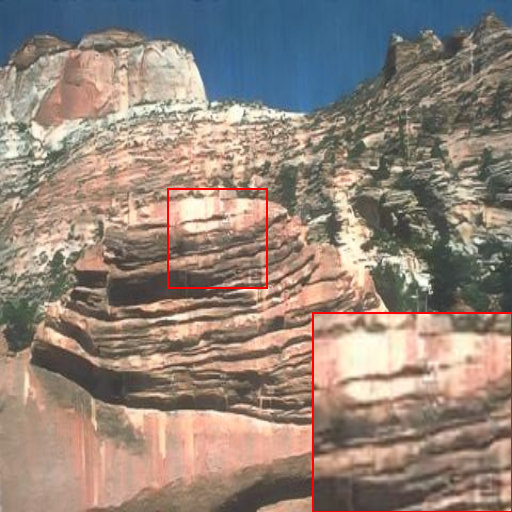}&\includegraphics[height=0.09\linewidth, width=0.15\linewidth]{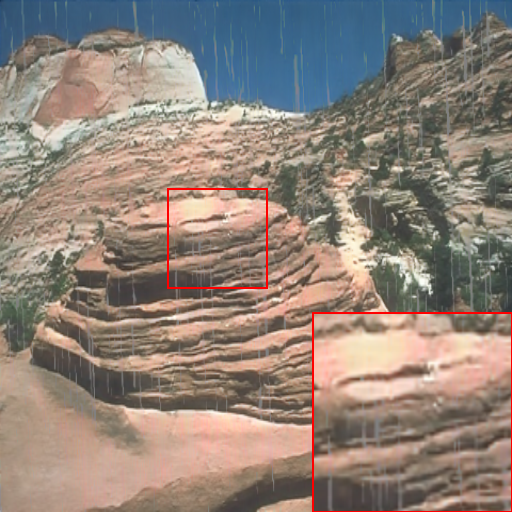}&\includegraphics[height=0.09\linewidth, width=0.15\linewidth]{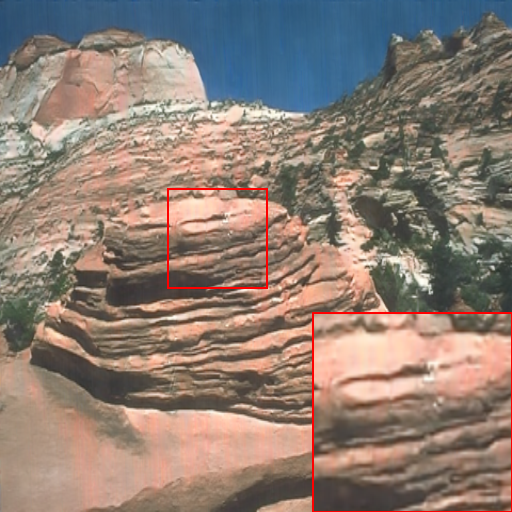}&\includegraphics[height=0.09\linewidth, width=0.15\linewidth]{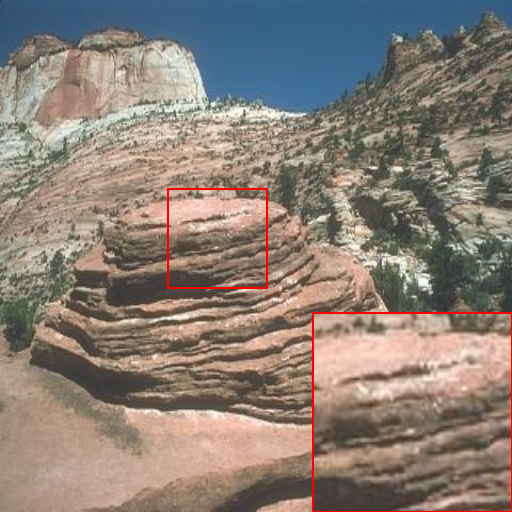}\\


        
        \rotatebox[origin=c]{90}{Rain100L\hspace{-40pt}}&\includegraphics[height=0.09\linewidth, width=0.15\linewidth]{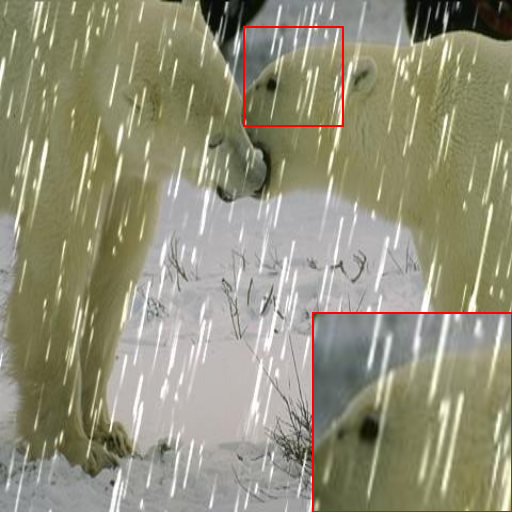}&\includegraphics[height=0.09\linewidth, width=0.15\linewidth]{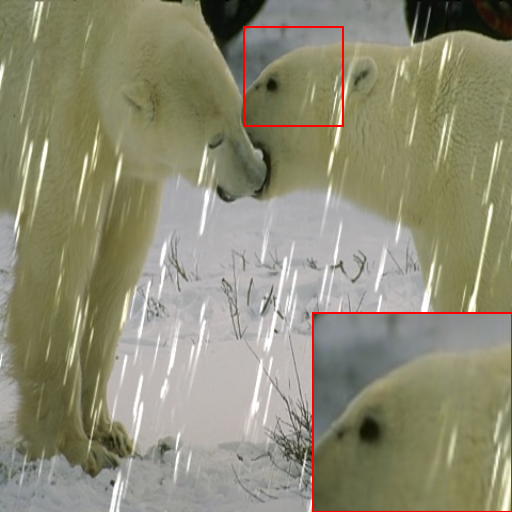}&\includegraphics[height=0.09\linewidth, width=0.15\linewidth]{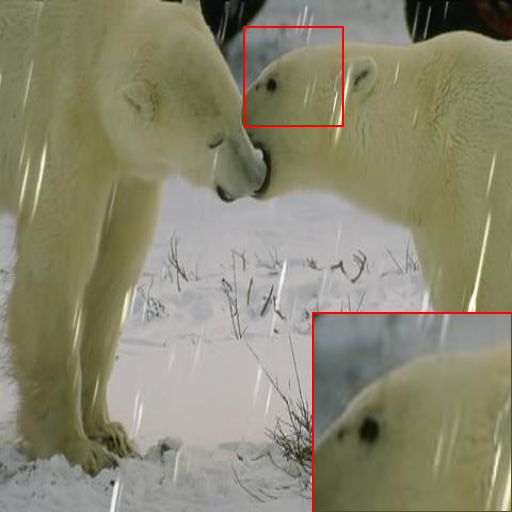}&\includegraphics[height=0.09\linewidth, width=0.15\linewidth]{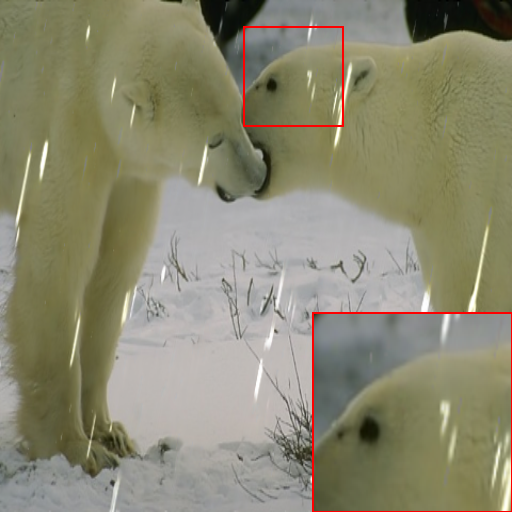}&\includegraphics[height=0.09\linewidth, width=0.15\linewidth]{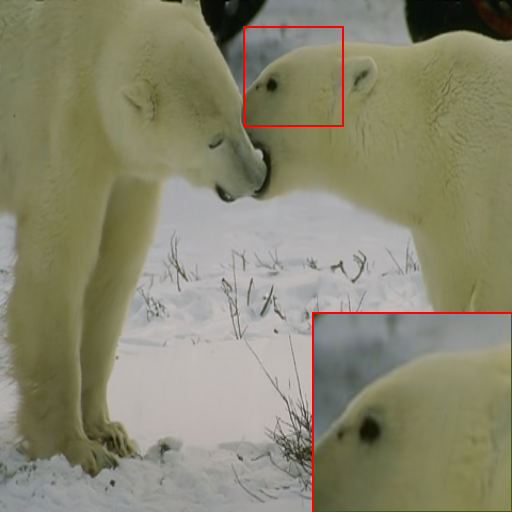}&\includegraphics[height=0.09\linewidth, width=0.15\linewidth]{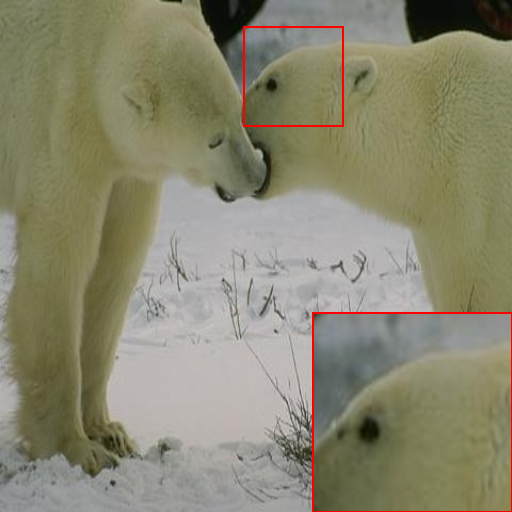}\\
        
        \rotatebox[origin=c]{90}{Snow100k\hspace{-40pt}}&\includegraphics[height=0.09\linewidth, width=0.15\linewidth]{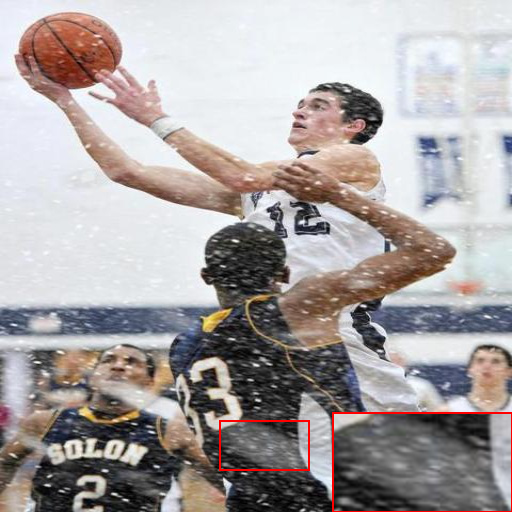}&\includegraphics[height=0.09\linewidth, width=0.15\linewidth]{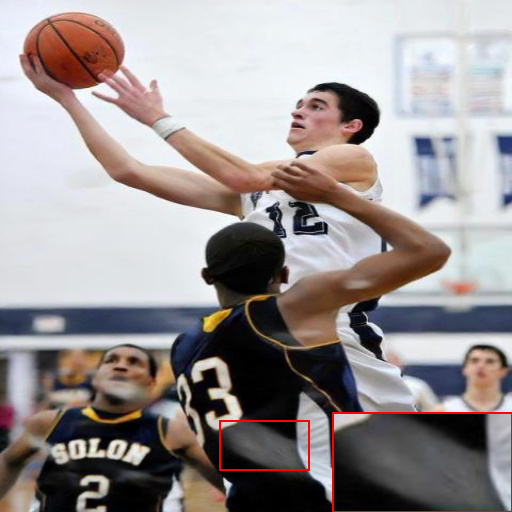}&\includegraphics[height=0.09\linewidth, width=0.15\linewidth]{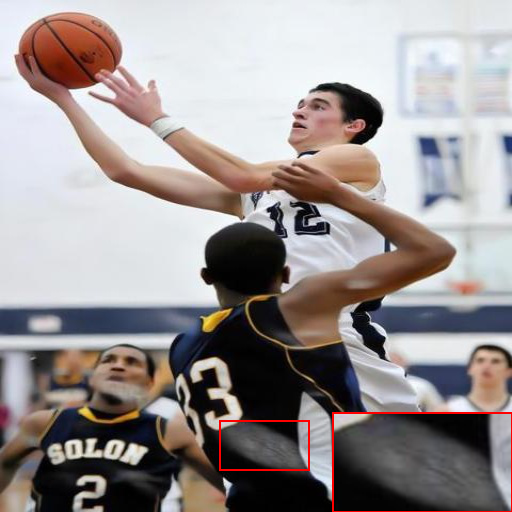}&\includegraphics[height=0.09\linewidth, width=0.15\linewidth]{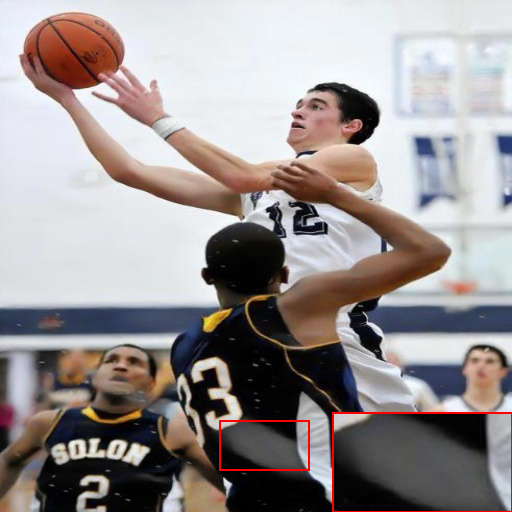}&\includegraphics[height=0.09\linewidth, width=0.15\linewidth]{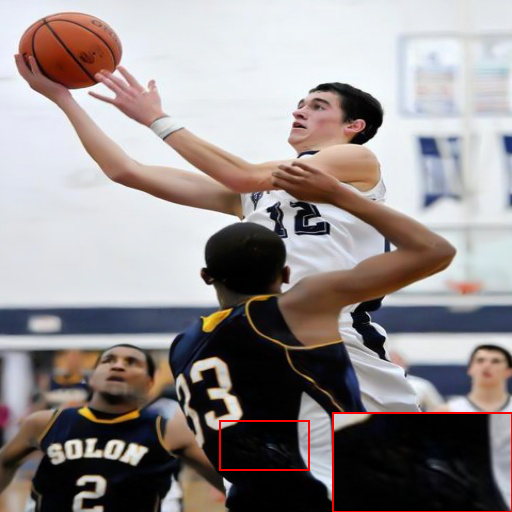}&\includegraphics[height=0.09\linewidth, width=0.15\linewidth]{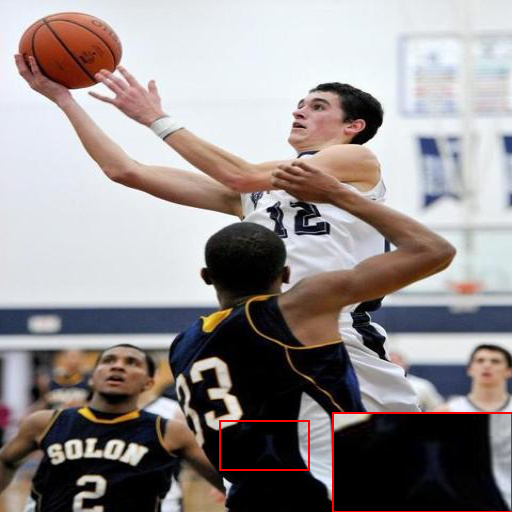} \\ 

        \end{tabular}
    \caption{Qualitative comparisons of AWRaCLe with top performing approaches (TSMC, PromptIR and DiffUIR) on SOTS, Rain100L, Rain100H and Snow100k datasets. Zoomed-in patches are provided for examining fine details.}
    \label{fig:qual}
\end{figure*}

\section{Ablation Studies}
\label{sec:ablations}
In this section, we first  demonstrate the effect of the context pair provided to AWRaCLe and Painter. We show that AWRaCLe uses degradation-specific information (DSI) from the context pair to guide restoration while Painter fails to use any DSI from the context. We then show the importance of the various components of AWRaCLe.


\newcommand{\blank}{{\mspace{0.7mu}\cdot\mspace{0.7mu}}}

\begin{table}[t]
    \centering
    \small
    \setlength{\tabcolsep}{0.8pt}
    \caption{Effect of fixing a specific context pair for the test sets described in Sec.~\ref{subsec:datasets} for different degradations.}
    \begin{tabular}{cccc}
        \toprule
        SOTS & Rain100L & Rain100H & Snow100k \\
        \midrule
        $31.61 \pm 0.18$ & $35.71 \pm 4\blank10^{-3}$ & $27.2 \pm 0.02$ & $33.48 \pm 0.01$ \\
        $0.981 \pm 2\blank10^{-4}$ & $0.966 \pm 10^{-4}$ & $0.84 \pm 3\blank10^{-4}$ & $0.934 \pm 3\blank10^{-4}$ \\
        \bottomrule
    \end{tabular}
    \label{tab:mean}
\end{table}


\subsection{Effect of Context Pairs}
\label{subsec:context}
We first analyze the performance of our method and the in-context learning method, Painter, for correct and incorrect context on the Rain100L dataset. As shown in Table~\ref{tab:quant_train}, when provided with correct context, AWRaCLe has a much higher PSNR (dB)/SSIM of $35.71/0.966$ compared to Painter ($23.19/0.900$). Next, we provided incorrect context, i.e., the degradation in the query image $I_\texttt{q}$ does not match the degradation present in $I_\texttt{d}$ of the context pair. Providing incorrect context to AWRaCLe yields a PSNR (dB)/SSIM of $25.93/0.826$ which is a $\sim10$ dB drop in performance with respect to correct context. However, Painter's values with incorrect context ($23.15/0.900$) are nearly unchanged from its performance with correct context, which indicates that it lacks utilization of the context for image restoration. Qualitative results for this experiment are in the supplementary. 



Next, we analyze the impact of specific context pairs on the performance of AWRaCLe. In Table~\ref{tab:mean}, we report mean ($\mu$) $\pm$ standard deviation ($\sigma$) of PSNR (row2) and SSIM (row3) obtained by randomly selecting $10$ paired context images for each of deraining, dehazing and desnowing, and fixing each of these context pairs over the entire test set. This is different from the testing strategy used in our experiments, where the context pair is randomly chosen for each image of the test sets. The table shows that AWRaCLe is quite robust to different context pairs from the same degradation. 

Finally, we tested our model’s robustness to out-of-distribution (OoD) context pairs by sampling them from the following datasets unseen during training: Foggy Cityscapes~\cite{cityfog} for dehazing, rain images from RainDS~\cite{rainds} for deraining, and SnowCityscapes~\cite{citysnow} for desnowing. Our model obtained PSNR/SSIM of $31.02/0.976$ on SOTS Outdoor, $35.72/0.966$ on Rain100L, $27.19/0.840$ on Rain100H and $33.40/0.934$ on Snow100k datasets. These results show only minimal deviations from those reported in Table~\ref{tab:quant_train}, showcasing our model’s resilience to out-of-distribution context.


\begin{table}[t]
    \small
    \centering
    \caption{Quantitative comparisons of different ablations conducted on AWRaCLe. All ablation settings are tested on the SOTS dataset. The best result is in bold.}
    \label{tab:ablations}
    \setlength{\tabcolsep}{3.5pt} 
    \begin{tabular}{lccc c}
        \toprule
        Context & DCE & CF & MLF & PSNR/SSIM \\
        \midrule
        -         & -         & -         & -         & 29.53/0.972 \\
        Paired    & -         & \checkmark & \checkmark & 30.53/0.978 \\
        Paired    & \checkmark & -         & \checkmark & 31.16/0.979 \\
        Paired    & \checkmark & \checkmark & -         & 30.12/0.977 \\
        Unpaired  & \checkmark & \checkmark & \checkmark & 29.84/0.974 \\
        Paired    & \checkmark & \checkmark & \checkmark & \textbf{31.65}/\textbf{0.981} \\
        \bottomrule
    \end{tabular}
\end{table}

\subsection{Effect of Individual Components}
\label{subsec:ablation}

In this section, we show the importance of the various components of AWRaCLe. 
Table~\ref{tab:ablations} shows quantitative results for each of our ablations on the SOTS dataset. In the table, ``Context'' refers to training with either paired or unpaired context, ``DCE'' indicates if the DCE block is used, ``CF'' indicates usage of CF block and ``MLF'' refers to the incorporation of multi-level fusion. A ``\checkmark'' in a column means that component is used, while a ``-'' means it is not used. The table shows that our proposed method, AWRaCLe (last row), demonstrates the best performance. 




\section{Conclusions}
We proposed a novel approach called AWRaCLe for all-weather image restoration that leverages visual in-context learning. We showed that suitably designed degradation context extraction and fusion blocks are central to the performance of our method. Additionally, we presented multi-level fusion of context information which is key to achieving good restoration performance. AWRaCLe advances the state-of-the-art in AWIR on standard datasets for the tasks of deraining, desnowing and dehazing. 
We believe that our method will be an important enabler for solving the complex AWIR task in its generality.

\section*{Acknowledgment}
This research is based upon work supported by the Office of the Director of National Intelligence
(ODNI), Intelligence Advanced Research Projects Activity (IARPA), via IARPA R\&D Contract
No. 140D0423C0076. The views and conclusions contained herein are those of the authors and
should not be interpreted as necessarily representing the official policies or endorsements, either
expressed or implied, of the ODNI, IARPA, or the U.S. Government. The U.S. Government is
authorized to reproduce and distribute reprints for Governmental purposes notwithstanding any
copyright annotation thereon.

\small
\bibliography{aaai25}

\end{document}